# Dispersion and Line Formation in Artificial Swarm Intelligence

DONGHWA JEONG and KIJU LEE, Case Western Reserve University

## 1. INTRODUCTION

One of the major motifs in collective or swarm intelligence is that, even though individuals follow simple rules, the resulting global behavior can be complex and intelligent. In artificial swarm systems, such as swarm robots, the goal is to use systems that are as simple and cheap as possible, deploy many of them, and coordinate them to conduct complex tasks that each individual cannot accomplish [Cao 1995; Simmons 2000; Sahin 2005]. The system may involve a group of artificially intelligent agents with limited sensing, communication, and computing capabilities and can achieve a higher level task by the agents locally or globally collaborating with each other. Local collaboration can be established among the neighbors within a certain communication range or physical/social distance. Global collaboration can be realized by centralized control possibly with a designated leader.

Global shape formation is one of the challenging problems in artificial swarm intelligence. In nature, it is performed for various purposes, such as engergy saving and path optimization. For instance, a flock of large birds fly together while forming a V shape in order to reduce the air resistance and fatigue for physically weak birds [Weimerskirch 2001]. Ants form a line to optimize the path to a food source by laying a trail of pheromone and regulate foraging activities without any central control or spatial information [Prabhakar 2012; Gordon 2013]. This trail enables the ant colony find the shortest path to the food source [Bonabeau 2000]. Shape formation in artificial intelligence systems is usually required for specific task-oriented performance, including 1) forming sensing grids [Spears 2004], 2) exploring and mapping in space, underwater, or hazardous environments [Nawaz 2010; Wang 2011], and 3) forming a barricade for surveillance or protecting an area/person [Cheng 2005]. This paper presents a dynamic model of an artificial swarm system based on a virtual spring damper model and algorithms for dispersion without a leader and line formation with an interim leader using only the distance estimation among the neighbors.

## 2. DYNAMIC MODEL OF COLLECTIVE SYSTEMS

Our dynamic model of swarm systems involving multiple agents is based on a virtual spring damper model. To realize attractive and repulsive forces in a stable manner, virtual damper is added to the spring system to depress the oscillatory motion of the agents. Modeling dynamic multi-agent systems frequently requires global optimization by designing objective functions. As the number of agents increases, the computational cost for optimization increases exponentially. To address this problem, we used geometric conditions that guarantee global shape formation from geometric primitives based on the attractive and repulsive forces between the agents as shown in Fig. 1. Each node and edge represent an agent and a communication/sensing/social connection respectively where the spring constant and damping ratio may be differently defined depending on each application. The mass of each node may indicate the weight or relative importance of that agent within the group.

Let $A_1, A_2, \cdots, A_n$ be the agents in two-dimensional space. For the $i^{th}$ agent, the two-dimensional Cartesian coordinate is denoted by $\vec{x}_i = [x_i, y_i]^T$. Likewise, the vector from the $i^{th}$ agent to the $j^{th}$ agent is given by $\vec{x}_{ij} = [x_i - x_j, y_i - y_j]^T$ where $||\vec{x}_{ij}|| = \sqrt{(x_i - x_j)^2 + (y_i - y_j)^2}$ Assuming that the $i^{th}$ agent has a mass, $m_i$. Then it follows the Newton's second law: $\vec{f}_i^m = m_i \ddot{\vec{x}}_i$ where $\vec{f}_i^m$ is the force at the $i^{th}$ node and $\vec{x}_i$ is its position. The force exerted by the spring and damper is described by $\vec{f}_i^k = -k_{ij}\vec{d}_{ij}$ and $\vec{f}_i^b = -b_{ij}\dot{\vec{x}}_{ij}$ where $k_{ij}$ is the spring constant, $b_{ij}$ is the damping coefficient between agent $i$ and neighboring agent $j$, and $\vec{d}_{ij}$ is a displacement vector between the two. In this model, each agent knows the distances to its neighbors only. However, $\vec{d}_{ij}$ is expressed as a vector since the sum of multiple interactions to an agent will force the agent to a certain direction in the global coordinate system. The net force acting on the $i^{th}$ agent in the global coordinate can be derived as

$$\vec{f}_i^m = \vec{f}_i^k + \vec{f}_i^b = -\sum_{j \in M_i}(k_{ij}\vec{d}_{ij} + b_{ij}\dot{\vec{x}}_{ij})$$

where $M_i$ is the number of neighboring agents of the $i^{th}$ agent. The acceleration is calculated as $\ddot{\vec{x}}_i = -\sum_{j \in M_i}(k_{ij}\vec{d}_{ij} + b_{ij}\dot{\vec{x}}_{ij})/m_i$. In practical control with a microprocessor, the above equation can be rewritten in discrete time representation given by

$$\dot{\vec{x}}_i[t+1] - \dot{\vec{x}}_i[t] = -\sum_{j \in M_i}(k_{ij}\vec{d}_{ij} + b_{ij}\dot{\vec{x}}_{ij}[t])/m_i.$$

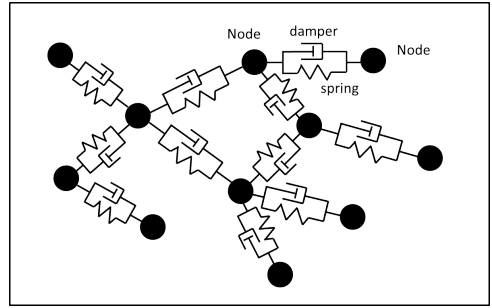

Fig. 1. Attractive and repulsive forces between the nodes are modeled with a spring and a damper.

In case of $m_i = 1$ for all $i = 1, \cdots, n$, the anticipated velocity and position at $t+1$ and $t+2$ are given by

$$\dot{\vec{x}}_i[t+1] = \dot{\vec{x}}[t] - \sum_{j \in M_i}(k_{ij}\vec{d}_{ij} + b_{ij}\dot{\vec{x}}_{ij}[t]); \quad \vec{x}_i[t+2] = 2\vec{x}_i[t+1] - \vec{x}_i[t] - \sum_{j \in M_i}(k_{ij}\vec{d}_{ij} + b_{ij}\dot{\vec{x}}_{ij}[t]).$$

3. DISPERSION AND LINE FORMATION

Algorithms

Two algorithms are introduced here: 1) dispersion based on trigonal plannar elements without a leader and 2) line formation using paired line elements using an interim leader.

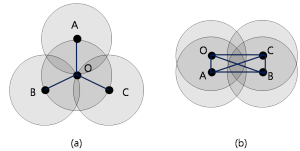

Fig. 2. Elementary geometric shapes: (a) trigonal planar and (b) paired line.

*Dispersion using trigonal planar elements.* Let $N$ be a set of dynamic agents. For every $O \in N$, label the closest three neighbor agents as $A, B$, and $C$. By using the proposed attractive and repulsive forces, let $L_{OA} = L_{OB} = L_{OC} = L_d$ where $L_d$ is the desired distance between the agents.

*Lining using paired line elements.* Let $N$ be a set of dynamic agents. For every $O \in N$, label the closest agent as $A$. Let $L_{OA} = \epsilon$ and form a pair, $(O, A)$. Find the nearest pair and label as $(B, C)$. Let $L_{OB} = L_{OC} = L_{AB} = L_{AC}$. Note that $L_{OB} = L_{OC}$ for $L_{OA} = L_{BC} = \epsilon$ as $\epsilon \to 0$. When all rectangulars are connected each other, two ends are designated as interim leaders of the connected chains in order to stretch the line to opposite directions.

Simulation Results

MATLAB simulation is performed with 50 agents randomly spread in two-dimensional space. The simple trigonal elements guarantee dispersion of the agents and paired line elements guarantees line shape while maintaining connectivity as shown in Fig. 3 and 4. If each agent has a limited sensing/communication range, the system may result in multiple disc formation or outliers. Fig. 5 shows the simulation results for two different communication ranges.

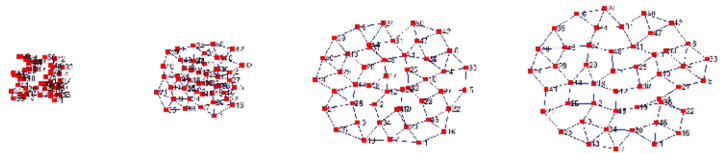

Fig. 3. Dispersion based on trigonal planar elements

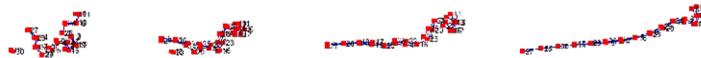

Fig. 4. Line formation using paired line elements

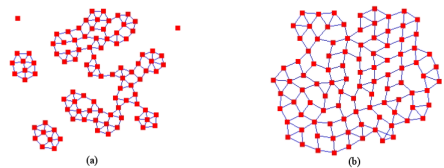

Fig. 5. Separated discs and connected disc with different sensing range: (a) 40 inch, (b) 60 inch with 100 agents